\documentclass[final,3p,times,twocolumn]{elsarticle}
 \usepackage{footnote}
\usepackage{caption}
\usepackage{booktabs}
\usepackage{flushend}
\usepackage[british]{babel}
 \usepackage{mathrsfs}
 \usepackage{amsmath}

\usepackage{caption}
\usepackage{tikz}
\usepackage{epsfig}
\usepackage{subfigure}
\usepackage{graphicx}
\usepackage{epstopdf}
\usepackage{multirow}
\usepackage{amssymb}
\usepackage{amsthm}
\usepackage{amsmath}

\usepackage{lineno}
\journal{ }

\begin{document}
\captionsetup[figure]{labelfont={bf},font={footnotesize},name={Fig.},labelsep=period}
\captionsetup[table]{labelfont={bf},font={footnotesize},name={Table},labelsep=period}
\begin{frontmatter}


\title{Optical Fringe Patterns Filtering Based on Multi-stage Convolution Neural Network}
\author[label1]{Bowen Lin}
\author[label1]{Shujun Fu\corref{cor2}}
\ead{shujunfu@163.com}
\author[label3]{Caiming Zhang}
\author[label2]{Fengling Wang}
\author[label4]{Yuliang Li}
\cortext[cor2]{Corresponding author.}
\address[label1]{School of Mathematics, Shandong University, Jinan 250100, China}
\address[label2]{College of Arts Management, Shandong University of Arts, Jinan 250300, China}
\address[label3]{School of Computer Science and Technology, Shandong University, Jinan 250101, China}
\address[label4]{Department of Intervention Medicine, The Second Hospital of Shandong University, Jinan 250033, China}
%
\begin{abstract}
Optical fringe patterns are often contaminated by speckle noise, making it difficult to accurately and robustly extract their phase fields.
To deal with this problem, we propose a filtering method based on deep learning, called optical fringe patterns denoising convolutional neural network (FPD-CNN),
for directly removing speckle from the input noisy fringe patterns. Regularization technology is integrated into the design of deep architecture. Specifically, the FPD-CNN method is divided into multiple stages, each stage consists of a set of convolutional layers along with batch normalization and leaky rectified linear unit (Leaky ReLU) activation function. The end-to-end joint training is carried out using the Euclidean loss. Extensive experiments on simulated and experimental optical fringe patterns, especially finer ones with high-density regions, show that the proposed method is competitive with some state-of-the-art denoising techniques in spatial or transform domains,efficiently preserving main features of fringe at a fairly fast speed.
\end{abstract}


\end{frontmatter}


\section{Introduction}
Optical interferometric techniques have been widely used in scientific research and engineering for its simple optical devices and the ability to
provide high resolution and full field measurements in a non-contact mode, such as electronic speckle pattern interferometry (ESPI).
Noise will unavoidably appear in the process of formation and acquisition of optical fringe patterns. In principle, a noisy fringe pattern can be modeled as
\begin{equation}
z\left ( i,j \right )=a\left ( i,j\right )+b\left ( i,j \right )cos\left ( \varphi \left ( i,j\right ) \right )+n\left ( i,j \right ),
\label{eq:1}
\end{equation}
where $\left ( i,j \right )$ is the spatial and temporal coordinate, $a\left ( i,j \right )$, $b\left ( i,j \right )$, $\varphi \left ( i,j \right )$ and $n\left ( i,j \right )$ are the background intensity, fringe amplitude, phase distribution and image noise, respectively \cite{1}$\sim$\cite{6}. The existence of speckle noise seriously affects the accurate phase extraction, which is very important for the success of fringe pattern demodulation. As the frequency spectrum of fringes and noise usually superimpose and cannot be separated clearly, simple filters have a strong blurring effect on fringe features, especially for fine ones with high density. Therefore, it is challenging to suppress the complicated speckle noise in the optical fringe pattern while preserving the features.

\begin{figure*}[t!]
  \captionsetup{margin=20pt,format=hang,justification=justified}
  \centering
  \includegraphics[width = 156 mm, height = 38 mm]{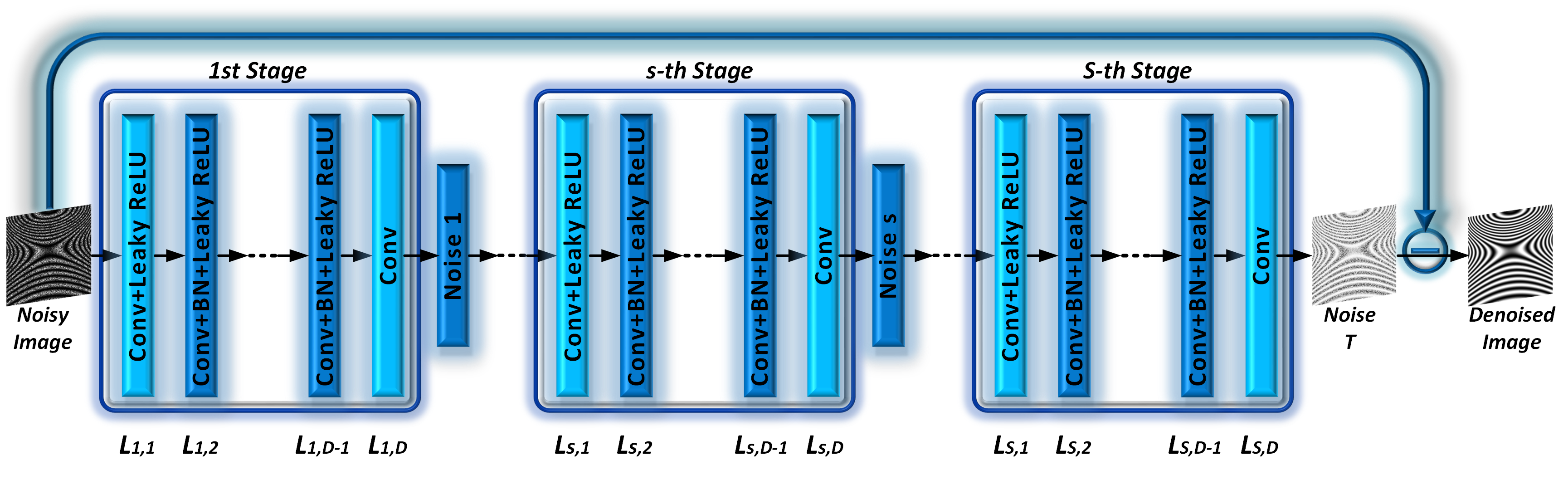}
  \caption{Network architecture of the proposed FPD-CNN.}
  \label{Fig:1}
\end{figure*}
Generally, various algorithms previously suggested for speckle noise reduction of optical fringe pattern can be categorized as methods in spatial domain or transform domain. For the transform domain techniques, Fourier transform \cite{7}, windowed Fourier transform (WFF) \cite{8}, and wavelet transform \cite{9}, have been applied with inspiring performance. For the spatial domain techniques, filtering along fringe orientation \cite{10} is an effective option, for example, Yu \emph{et al.} proposed spin filters \cite{11} to smooth fringe patterns along the local tangent orientation. Tang \emph{et al.} proposed a technique using second-order oriented partial differential equations for diffusion along the fringe orientation \cite{12}. Wang \emph{et al.} proposed a coherence enhancing diffusion (CED) for optical fringe pattern denoising \cite{13}. This technique controls the smoothing both along and perpendicular to the fringe orientation. This principle is later developed for phase map denoising \cite{14}. However, Fu \emph{et al.} \cite{15} developed a nonlocal self-similarity filter, which averages the intensity of similar pixels searched in whole image instead of in a local fringe direction as the spin filters do. Involving more pixels with higher similarity levels and being free of the fringe orientation estimation, this simple algorithm has stronger robustness against noise with better filtering results, especially in the processing of fine optical fringe pattern. The variational image decomposition technique proposed in the past few years has been successfully applied to image denoising, such as ones in \cite{16}$\sim$\cite{21}. The basic idea of applying these methods for fringe pattern denoising is to decompose the image into low-density fringes, high-density fringes and noise parts, each of which is modeled by a different functional space. To this end, an optimization functional is established by combining the corresponding norms in uncorrelated space and each part is filtered by minimizing this functional. Recently, a denoising method \cite{22} for discontinuous fringe patterns with joint fuzzy c-means (FCM) clustering and partial differential equations is proposed. Firstly, discontinuous regions in fringe images are identified, and then denoised by an adaptive shape-preserving oriented partial differential equation model with the controlling speed function. The noise is eliminated while the shape of fringes and the discontinuity are kept.

However, when facing fine optical fringe patterns contaminated by complicated speckle noise (such as fringes with the width of only few pixels in some high-density/high-frequency regions, and even the case of no more than 5 pixels), most existing methods face the difficulty of making trade-offs between good filter performance and time cost, especially when parameter adjustments and algorithm execution. Although the method based on the diffusion equation in the spatial domain is faster, it will produce severe blurring effects in the fine high-density region, such as the CED method \cite{13} for its inaccurate estimation of the fringe orientation. However, the estimation of fringe orientation is not a trivial work, particularly for high-density regions whose fringe structure was destroyed by strong speckle noise, and parameter tuning is extremely complex. Another important issue is that the geometry of the fringe on boundary is easily distorted. The methods in transform domain often have high computational cost and produce artifacts easily, such as the WFF method \cite{8}.

A more recent trend is deep learning \cite{23}, which has arisen as a promising framework providing state-of-the-art performance for image classification \cite{24,25} and segmentation \cite{26}$\sim$\cite{28}. Moreover, regression-type neural networks have demonstrated impressive results on inverse problems with exact models such as image denoising \cite{29}$\sim$\cite{34}, deconvolution \cite{35}, artifact reduction \cite{36,37}, super resolution (interpolation) \cite{38,39}. Central to this resurgence of neural networks has been the convolutional neural network (CNN) architecture. Deep learning as a powerful data processing technology has also penetrated into the field of optics, such as phase recovery and holographic image reconstruction \cite{40,41}, identification and classification of objects hidden behind scattering media \cite{42}, coherent noise reduction in three-dimensional quantitative phase imaging \cite{43}.

To address the reduction of speckle noise for finer fringe pattern with high-density regions,  we proposed an optical fringe pattern denoising convolutional neural network model (FPD-CNN) based on deep learning. The design of the architecture according to the derivation of regularization theory.  Especially the residual learning technique \cite{44} is introduced into the network model by the solution of regularization model. The architecture of the model is divided into multiple stages, each of which consists of several convolutional layers along with batch normalization \cite{45} and Leaky ReLU \cite{46} activation function (see Fig.\ref{Fig:1}). We also verified the importance of the activation function leaky rectified linear unit adopted in the proposed network. It is trained in an end-to-end fashion using the Euclidean loss function. One of the main reasons of using deep-learning-based techniques for fringe denoising is that they learn parameters for  image restoration  directly from the training data rather than relying on predefined image priors or filters. Therefore, the proposed method can be regarded as an adaptive method that relies on large-scale data. The advantage of the proposed method is that FPD-CNN model not only has better denoising performance but also has better performance than the compared algorithm in preserving boundary features of the fringe pattern, especially the fine ones. Most importantly, the proposed FPD-CNN has a faster denoising speed, which is verified by experiments.

The remainder of the paper is organized as follows. Section 2 briefly introduces the notions and preliminaries of fringe pattern modeling. Section 3 presents the proposed denoising CNN model in details and the related mathematical principle. Section 4 shows experimental results of our method by comparing it with other methods to validate its effectiveness. Section 5 concludes the paper.
\section{ Notions and Preliminaries of Fringe Pattern Modeling }
In this section, we briefly introduce the modeling of fringe pattern. Optical interference produces fringe patterns \cite{3}. Given the object beam $A_{0}( i,j; t )$ and the reference beam $A_{r}(i,j;t)$:
\begin{equation}
A_{0}(i,j;t)=a_{0}(i,j;t)e^{\textsl{\textbf{j}}(-\omega t+\varepsilon+\varphi_{0}(i,j;t))},
\label{eq:2}
\end{equation}
\begin{equation}
A_{r}(i,j;t)=a_{r}(i,j)e^{\textsl{\textbf{j}}(-\omega t+\varepsilon+\varphi_{r}(i,j))},
\label{eq:3}
\end{equation}
where $(i,j)$ and $t$ are spatial and temporal coordinates, $a_{0}(i,j;t)$ and $a_{r}(i,j)$ are the amplitude, and $-\omega t$, $\varepsilon$, $\varphi_{0}(i,j;t)$
and $\varphi_{r}(i,j)$  are phase terms of the corresponding beam. $\omega$ is the angular frequency and $\varepsilon$ is an initial phase. $\varphi_{0}(i,j;t)$ reflects optical path of the object beam, while the $\varphi_{r}(i,j)$ reflects optical path of the reference beam that is usually time-invariant. If the object beam and the reference beam vibrate in the same direction and superpose on the detector, the resulted optical intensity is
\begin{equation}
I(i,j;t)=\left | A_{0}(i,j;t)+A_{r}(i,j;t) \right |^{2},
\label{eq:4}
\end{equation}
and we can derive that
\begin{equation}
 \begin{split}
I(i,j;t)&=a_{0}^{2}(i,j;t)+a_{r}^{2}(i,j;t)+2a_{0}(i,j;t)a_{r}(i,j)\\
&\times cos(\varphi_{0}(i,j;t)-\varphi_{r}(i,j)).
\end{split}
\label{eq:5}
\end{equation}
For two correlated speckle fields at time instances $t_{0}$ and $t$, a speckle correlation fringe pattern is defined as
\begin{equation}
I(i,j;t)=\left | I(i,j;t)-I(i,j;t_{0}) \right |^{2}.
\label{eq:6}
\end{equation}
Assuming that $a_{0}(i,j;t)=a_{0}(i,j;t_{0})$, Eq.(\ref{eq:6}) can be derived as
\begin{equation}
\begin{split}
I(i,j;t)&= 4a_{0}^{2}a_{r}^{2}+4a_{0}^{2}a_{r}^{2}cos(\Delta\varphi_{0}(i,j;t_{0},t)+\pi)\\
&+n(i,j;t_{0},t),
\end{split}
\label{eq:7}
\end{equation}
which include the noise term
\begin{equation}
\begin{split}
n(i,j;t_{0},t)&=-4a_{0}^{2}a_{r}^{2}(1-cos(\Delta\varphi_{0}(i,j;t_{0},t)))\times\\
&cos(\varphi_{0}(i,j;t_{0})+\varphi_{0}(i,j;t)-2\varphi_{r}(i,j)),
\end{split}
\label{eq:8}
\end{equation}
where the $\Delta\varphi_{0}(x,y;t_{0},t)$ is the phase difference between these two time instances $t_{0}$ and $t$, i.e
\begin{equation}
\Delta\varphi_{0}(i,j;t_{0},t)=\varphi_{0}(i,j;t)-\varphi_{0}(i,j;t_{0}),
\label{eq:9}
\end{equation}
which is a regular function and usually spatial (piecewise) smooth. $I(x,y;t)$ becomes a waving structure,
namely, a fringe pattern. The corresponding optical technique is electric speckle pattern interferometry. Both fringe patterns in Eq.(\ref{eq:5})
and Eq.(\ref{eq:7}) can be generally and mathematically modeled as Eq.(\ref{eq:1}), which is also suitable for fringe patterns formed by many other modalities.

\section{\textbf{Proposed Method}}
To make it easier to understand the proposed method, we first introduce the details of the network architecture, and then introduce the mathematical principle of the architecture design.
\subsection{Architecture of FPD-CNN}
For the design of our network architecture, the FPD-CNN method catains \emph{S} stages of noise estimation in the network and each stage has $D$
layers $(L_{s,d}, s=1,\ldots,S; d=1,\ldots,D)$ that means the proposed architecture is deeper. Then the denoisied fringe is obtained by simply subtracting the estimated
noise from noisy fringe pattern. Because the noise estimated by single stage  is not accurate, and often still contains some structural details of the fringe, a multi-stage
framework scheme is adopted to further reduce the error, which is similar to iterative regularization. The estimated noise at each stage serves as the input for the next stage.
As shown in Fig.\ref{Fig:1}, the architecture of each stage \emph{s} is described as follows:\\
\vspace{0.1em}\quad (i) Layer $L_{s,1}$(light blue layer) includes the spatial template convolution filtering (Conv) and the Leaky ReLU operations, where 64 filters of
size $5\times5\times1$ are used to generate 64 feature maps, and the Leaky ReLU is then utilized for nonlinearity. Here 1 indicates that the data to be processed is
one-dimensional.\\
\vspace{0.1em}\quad (ii) Layers $L_{s,2}$$\sim$$L_{s,D-1}$ (blue layers) includes the convolution, the batch normalization (BN) and the Leaky ReLU operations,
where 64 filters of $5\times5\times64$ are used, and the batch normalization is added to the middle of the convolution and the Leaky ReLU.\\
\vspace{0.1em}\quad (iii) Layer $L_{s,D}$ (light blue layer) includes only the convolution operation, where single filter of size $5\times5\times64$ is used to
reconstruct the output.

Furthermore, we pad zeros before convolution to ensure that each feature map of the middle layers has the same dimension as the noisy input. The batch normalization
is added to alleviate the internal covariate shift by incorporating a normalization step and a scale and shift step before the Leaky ReLU operation in each layer. Finally,
the Leaky ReLU function is defined as
\begin{equation}
h=\max\{z,0\}+\alpha\min\{z,0\},
\label{eq:10}
\end{equation}
which is a pixel by pixel operation. When $\alpha=0$, it will degenerate into the ReLU function. The Leaky ReLU activation function inherits the advantages of ReLU in two
aspects \cite{46}: the first is to effectively solve the so called ``exploding/vanishing gradient"; the second is to accelerate the convergence speed. At the same time,
it can propagate negative gradients forward which can reduce the risk of falling into local optimum.

\subsection{Mathematical Principle of FPD-CNN}
 The purpose of training the network in each stage is actually to solve the following regularization model as mentioned in \cite{30,31}:
\begin{equation}
\underset{x}{min}~E\left(x\right)=\frac{1}{2}\left \| z-x \right \|_{F}^{2}+\lambda \sum_{m=1}^{M}\sum_{n=1}^{N}\rho _{m}\left ( \left ( f_{m}\ast x \right )_{n} \right ).
\label{eq:11}
\end{equation}
Here, $z$ is a noisy observation, $x$ is the ground truth fringe image, $\left\|\cdot \right \|_{F}$ is the Frobenius norm, $N$ denotes the image size, $\lambda$ is a
regularization parameter, $M$ is the total number of all filter kernels, $f_{m}\ast x$ stands for the convolution of the image $x$ with the $m$-$th$ filter kernel, and $\rho_{m}(\cdot)$ represents the $m$-$th$ penalty
function. The solution of Eq.(\ref{eq:11}) can be interpreted as performing one gradient descent process at the starting point $z$, given by
\begin{equation}
\tilde{x}=z-\frac{\tilde{\alpha}}{\lambda}\sum_{m=1}^{M}\left ( f_{m}^{*}\ast \phi  _{m} \left ( f_{m}\ast z \right )\right ),
\label{eq:12}
\end{equation}
where $f_{m}^{*}$ is the adjoint filter of $f_{m}$ (i.e., $f_{m}^{*}$ is obtained by rotating the filter $f_{m}$ 180 degrees), $\tilde{\alpha}$ is a step size
and ${\phi}_{m}={\rho}'_{m}$ (derivative of $\rho$ with respect to $m$, the influence function ${\phi}_{m}$ can be regarded as pointwise nonlinearity operator
applied to convolution feature maps), replaced by Leaky ReLU activation function in our model. Eq.(\ref{eq:12}) is equivalent to the following formula:
\begin{equation}
v=z-\tilde{x}=\frac{\tilde{\alpha}}{\lambda}\sum_{m=1}^{M}\left ( f_{m}^{*}\ast \phi  _{m} \left ( f_{m}\ast z \right )\right ),
\label{eq:13}
\end{equation}
where $v$ is the estimated residual of $x$ with respect to $z$. Hence, we adopt the residual learning formulation to train a residual mapping defined as
\begin{equation}
\Re\left ( z,\Theta \right )=v,
\label{eq:14}
\end{equation}
where $\Theta$ is the network parameters, including filters and batch normalization parameters in each layer which are updated with data training. The residual $v$ is regarded as the noise estimated from $z$, and the approximate clean fringe image can be obtained by
\begin{equation}
\tilde{x}=z-\Re \left ( z,\Theta \right ).
\label{eq:15}
\end{equation}

Suppose that $Z=\left \{ \left ( x_{k},z_{k} \right ) \right \}_{k=1}^{K}$ represents $K$ clean-noisy training image pairs, and define the Euclidean loss function for each pixel as
\begin{equation}
\pounds \left ( \Theta  \right )=\frac{1}{2K}\sum_{k=1}^{K}\left \| \Re \left ( z_{k},\Theta  \right )-\left ( z_{k}-x_{k} \right ) \right \|_{F}^{2}.
\label{eq:16}
\end{equation}
Therefore, the training task for learning parameters $ \Theta$ can be considered as solving the optimization problem:
\begin{equation}
\left\{\begin{matrix}
&\underset{\Theta}{min}~~~\pounds \left ( \Theta  \right )=\frac{1}{2K}\sum_{k=1}^{K}\left \| \Re \left ( z_{k},\Theta  \right )-\left ( z_{k}-x_{k} \right ) \right \|_{F}^{2}\\
&s.t\quad v_{k}=\Re \left ( z_{k},\Theta  \right )=\frac{\tilde{\alpha}}{\lambda}\sum_{m=1}^{M}\left ( f_{m}^{*}\ast \phi  _{m} \left ( f_{m}\ast z_{k} \right )\right ).\\
&k=1,\cdots ,K
\end{matrix}\right.
\label{eq:17}
\end{equation}
Eq.(\ref{eq:17}) can be regarded as a two-layer feed-forward CNN. We apply the back-propagation algorithm to solve the problem (Eq.(\ref{eq:17})). Based on the theory of numerical approximation, our ultimate goal is essentially to train a highly nonlinear function (i.e.,$\Re$) as an approximation of the intensity statistical distribution of noisy pixels.

It is well known that increasing the number of layers appropriately can improve the performance, and we naturally add a multi-stage framework to the design of network architecture. Gradient descent algorithm optimizes the parameters $\Theta_{s}, s=1...S$ of the \emph{s-th} stage, so as to minimize the loss function $\pounds \left ( \Theta_{s}  \right )$. Generally, the training task first randomly divides the total data $\left \{ \left ( x_{k},z_{k} \right ) \right \}_{k=1}^{K}$ into several mini-batches \cite{44}, each mini-batch denoted as $Z_{1...P}=\left \{ \left ( x_{p_{q}},z_{p_{q}} \right ) \right \}_{p_{q}=1}^{P}$ of size \emph{P}, $q$ represents the\emph{ q-th } mini-batch. And the multi-stage
\begin{figure*}[ht]
\centering
\includegraphics[width=14.5cm,height=3.5cm]{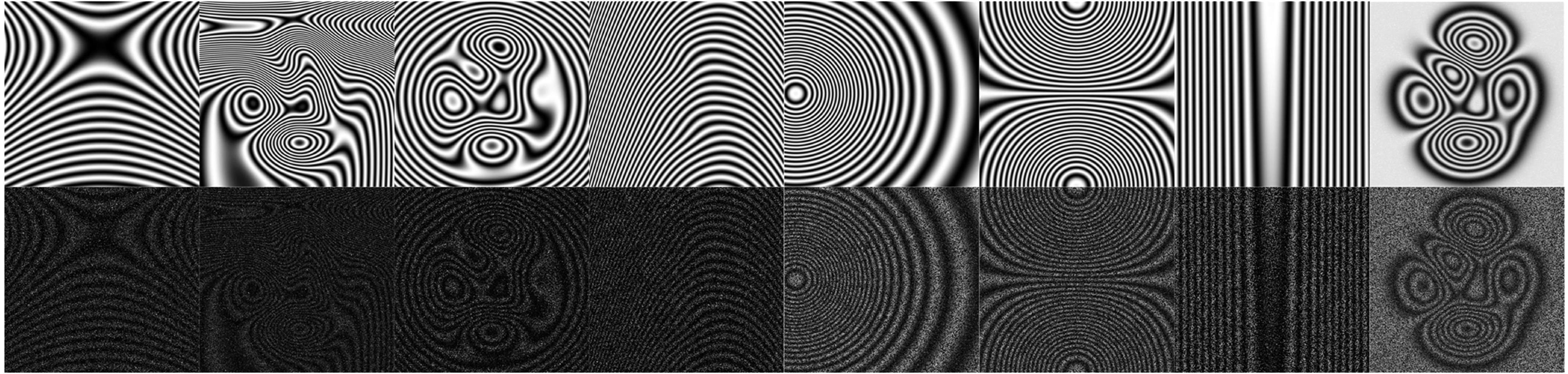} 
\caption{Partial image pairs simulated for training.}
\label{Fig:2}
\end{figure*}
\begin{figure*}[ht]
\centering
\includegraphics[width=14.5cm,height=3.5cm]{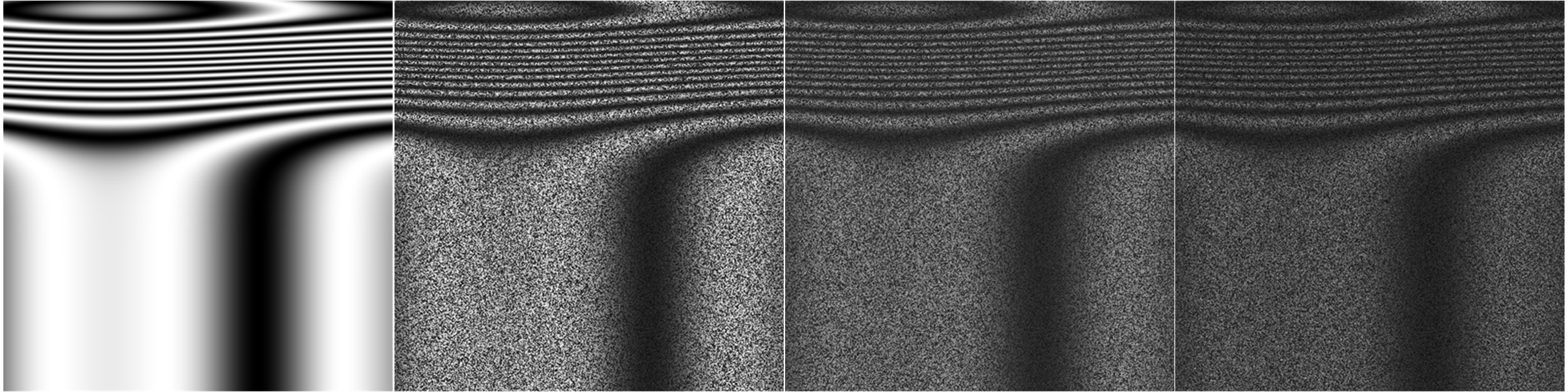} 
\caption{From left to right:  noise free image, three noisy images corresponding to $\lambda=$ 0, 5 and 10, respectively.}
\label{Fig:3}
\end{figure*}
framework is implemented on each mini-batch sequentially based on the end-to-end joint training fashion. The mini-batch is used to approximate the gradient of the loss function with respect to the parameters, by computing $\frac{1}{P}\frac{\partial\pounds(z_{p_{q}},\Theta_{s})}{\partial\Theta_{s}}$. With Gradient descent algorithm, the training proceeds in steps, and at each step we consider a mini-batch, so it's a total of $Q=K/P$ iterations for each epoch.
\section{Experimental Results With Analysis}
In this section, we first introduce the network training, including training and testing data, brief introduction of parameter setting and training implementation,
and then we present the results of our proposed FPD-CNN algorithm on both synthetic and the experimentally obtained fringe patterns.
We compare the proposed method with two representative methods, including WFF \cite{8} and CED \cite{13}, both of which belong to transform domain and spatial domain,
respectively.  In order to quantitatively evaluate and analysis the denoising performance of the proposed method, on the simulation data we will calculate the following
three image quality metrics: the peak signal to noise ratio (PSNR), the mean of structural similarity index (SSIM) \cite{47}, the mean absolute error (MAE).
For on the experimentally obtained data as we cannot get the clean reference, so we compare the performance of these methods through visual subjective evaluation.
The skeletons are also obtained by parallel thinning algorithm \cite{48} for the two types of data. Here, the comparison of the quality of the extracted corresponding
skeleton lines also reveal the performance of the algorithm. Another important evaluation indicator is to compare the execution time of methods.
\subsection{ Network Training}
\subsubsection{ Training and Testing Data}
In order to train our model, we use Eq.(\ref{eq:7}) to generate training and testing dataset.
For the phase function $\Delta\varphi_{0}(i,j;t_{0},t)$ in Eq.(\ref{eq:7}) are adopted different smooth (piecewise) functions, as shown in the following formula:
\begin{equation}
\Delta\varphi_{0}(i,j)=\sum_{l=1}^{L}\kappa_{l}\chi_{l}\left ( i,j \right ),
\label{eq:18}
\end{equation}
where $L$ represents the number of all compound functions, $\chi_{l}\left ( i,j \right )$ are set as compound function in exponential form,
polynomials (such as quadratic polynomial) or the product of compound function in exponential form and polynomials. The combined coefficients $\kappa_{l}$ take some random values. Its form is similar to the function : $\Delta\varphi_{0}(i,j)=55e^{-\frac{\left ( 2i-400 \right )^2}{20000}-\frac{\left ( 2j-162 \right )^2}{25000}}+2\left ( \frac{\left ( i-255 \right )^2}{400}+\frac{\left ( j-5 \right )^2}{700}  \right )e^{\frac{\left ( i-10\right )^2}{950000}+\frac{\left( j\right )^2}{65000}}+\frac{\left ( 2i-155 \right )^2}{300}+\frac{\left ( j-45 \right )^2}{200} $.
We refer to the simulation process in references \cite{3,4,10,13}. The other parameter settings in Eq.(\ref{eq:7}) are as follows for generating clean-noisy image pairs. To generate clean images $a_{0}^{2}$ is randomly taken in the range of [1, 150], denoted as $a_{0c}^{2}$ in Eq.(\ref{eq:7}) not including noise term $n$. In the noisy images, $a_{0}^{2}=a_{0c}^{2}+NED(\lambda)$, where $NED$ is a random variable subject to the negative exponential distribution, its expectation $\lambda$ is randomly taken in the range of $[0,50]$. $\varphi_{0}$ is uniformly distributed in $(-\pi,\pi]$, $\varphi_{r}=0$ and $a_{r}^{2}=1$. In the noisy image, $a_{0}^{2}$ is actually corrupted by $NED$, and this imperfection is shifted into the noise term $n$ \cite{3,4}.

1700 clean-noisy image pairs with the size of $256 \times 256$ are simulated. In this article, we standardize linearly the intensity range of image pixels to $\left [0, 255 \right]$. We add additive white Gaussian noise with mean 0 and standard deviation 10 to the image of the randomly selected 500 clean image for data augmentation. Then, 100 pairs are randomly selected for testing, and the rest for training. Partial image pairs are shown in Fig.\ref{Fig:2}. We believe that the reason for low contrast is that speckle noise reduces the mean level of image pixel intensity. In the simulation process, we also observe that the contrast of the noisy image will decrease as the expectation $\lambda$ increases under the condition that other variables in Eq.(\ref{eq:7}) are fixed. For example, we set these variables as follows, $a_{0c}^{2}=45$, $\Delta\varphi_{0}(i,j)=10e^{-\frac{\left ( i-110 \right )^{2}}{50000}}+180e^{-\frac{\left ( j-10 \right )^{2}}{50000}}-\pi$, $\lambda$ is set to 0, 5, and 10 to obtain three noisy images with size of $400\times400$ as shown in Fig.\ref{Fig:3}. The means of pixel intensity in 100th column of four images in Fig.\ref{Fig:3} are 185.33, 97.37, 53.18, and 34.25 respectively, which explains the reason of causing low contrast. Our aim is to put the simulated samples with different contrast into the training set to enhance network model's performance in denoising the experimentally obtained fringe patterns.

To avoid over fitting to some extent, the common data augmentation is also used, such as horizontal inversion, rotation transformation and other geometric transformation methods. Finally, 230400 clean-noisy patch pairs with patch size of $80 \times 80$ are cropped to train the proposed CNN model.
\begin{figure}[ht]
\centering
\includegraphics[width=5.8cm,height=7.1cm]{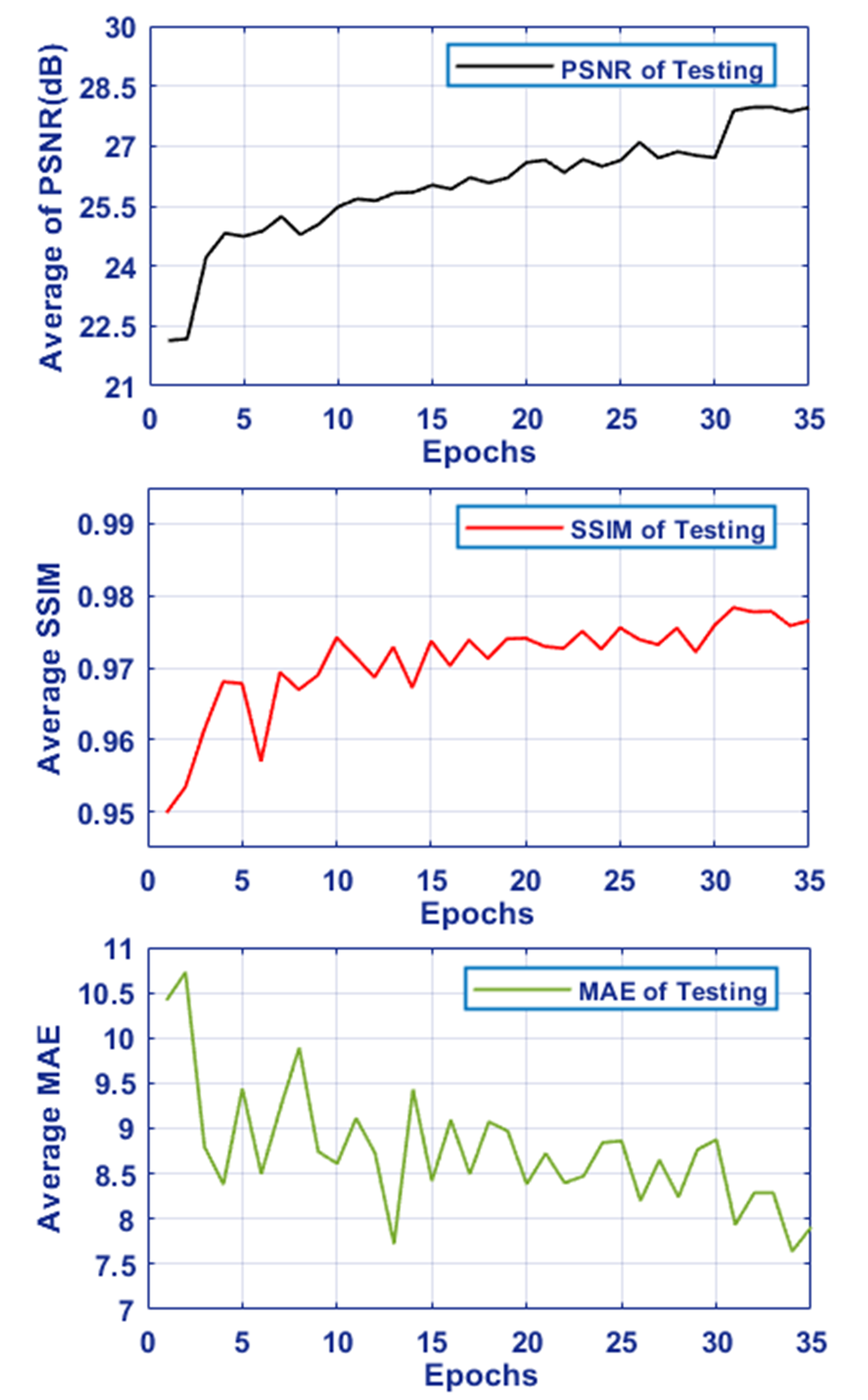} 
\caption{The change trend of average PSNR, average SSIM and average MAE on testing data.}
\label{Fig:4}
\end{figure}
\subsubsection{ Parameter Setting and Training}
The effective eight-layer architecture of convolution neural network is adopted in \cite{24,33}. In particular, Wang \emph{et al.} proposed an eight-layer network model to remove speckle noise in SAR, which has achieved advanced performance in both quantitative indicators and visual evaluation \cite{33}. But the noise model of fringe pattern used here is different from that of SAR image, which is commonly modeled as multiplicative noise model. Based on the simulation process of real imaging in section 4.1.1, the noise level of simulated noisy images is unfixed due to the randomness of parameters. In order to capture enough spatial feature information in feature abstraction for image denoising, we set \emph{S} and \emph{D} to be 3 and 8 respectively. The network depth has actually reached a relatively deeper 24 layers. Furthermore, the multi-stage framework increases the ease of use of the network model by simply adjusting the number of stages \emph{S} in order to achieve better performance.

ReLU is actually a piecewise linear function which prunes the negative part to zero, and retains the positive part. This results in a desirable sparsity, which is commonly believed to help ReLU achieve superior performance \cite{46}, such as finding better minima and promoting convergence. But this also leads to its inability to propagate negative gradient in parameter updating that easily causes the so-called ``neuronal death" during training. Here, we set the slope $\alpha$ of the first convolutional layer of each stage (that is the light blue layer in Fig.1, namely the $L_{s,1}$-\emph{th} layer) to be relatively small, so that it is close to the performance of ReLU, which promotes sparsity; in the subsequent convolutional layers, we set Leaky ReLU's slope $\alpha$ slightly larger than that of the first layer to enhance the propagation of negative gradient during training, thereby the learning parameters of the network can be constantly updated. In the experiments, we set the slope $\alpha$ to 0.05 for the $L_{s,1}$-\emph{th} layer of each stage, and the remaining layers $\alpha$ to 0.5.

The weights of all filters are initialized by the method in \cite{49}, and the whole network is trained using the optimization method of the adaptive moment
estimation (ADAM) \cite{50}. With a batch size of 64 and a learning rate of 1e-3. We just trained 35 epochs for our FPD-CNN. Fig.\ref{Fig:4} shows the change trend of average PSNR, average SSIM and average MAE for all training  epochs on the testing data.
Experiments are carried out in MATLAB R2017b using MatConvNet toolbox, with an Intel Xeon E5-2670 CPU 2.6GHz, and an Nvidia GTX1080 Founders Edition GPU. It takes about 40 hours on GPU.
\subsection{Ablation Study}
The purpose of this section is to illustrate why Leaky ReLU is used in the proposed CNN model compared to the well-known ReLU under the limited training data set.
\begin{figure}[!h]
\centering
\includegraphics[width=7.6cm,height=2.5cm]{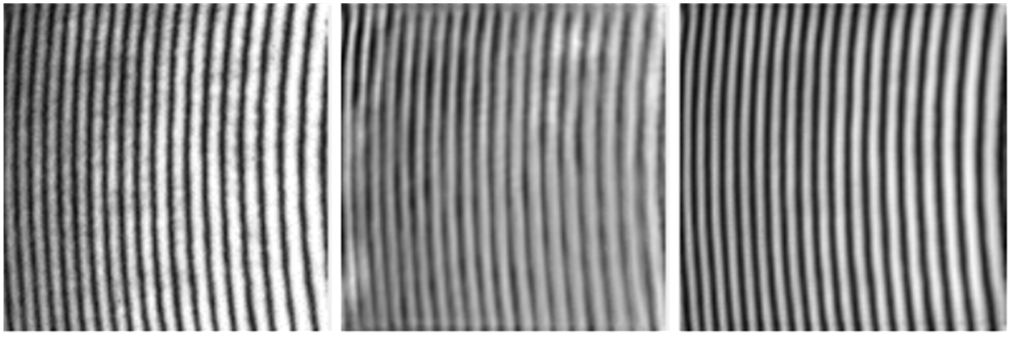}
\caption{Comparison of different activation functions (from left to right): the experimentally obtained optical fringe pattern, results of proposed FPD-CNN using ReLU and Leaky ReLU, respectively.}
\label{Fig:5}
\end{figure}
Therefore, An ablation study is performed to demonstrate the effects of Leaky ReLU in comparison with ReLU. Here, the same training data is used to train the proposed network, and the other configuration parameters are also the same.
\begin{figure*}[t]
\centering
\includegraphics[width=16.1cm,height=6.1cm]{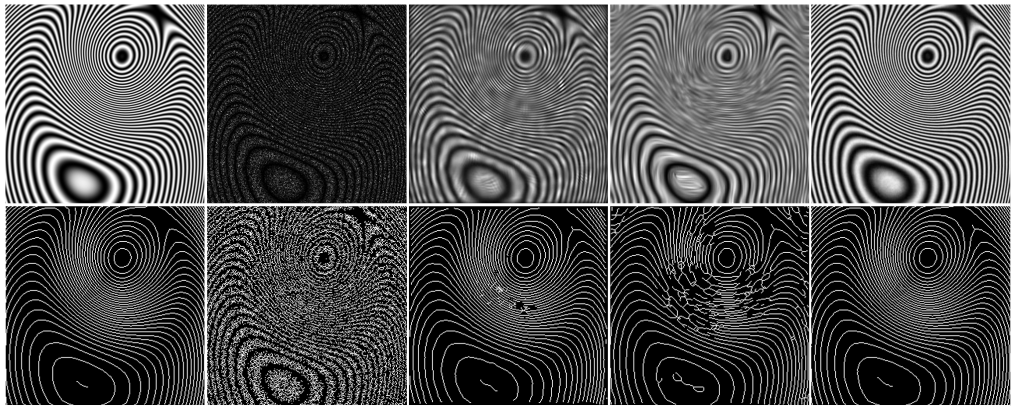}
\caption{Denoising a fine simulated fringe pattern with high-density regions (the 1st row): noise free image, noisy image, WFF, CED, FPD-CNN, respectively. The 2nd row are the corresponding skeletons.}
\label{Fig:6}
\end{figure*}
\begin{figure*}[t]
\centering
\includegraphics[width=16.1cm,height=6.1cm]{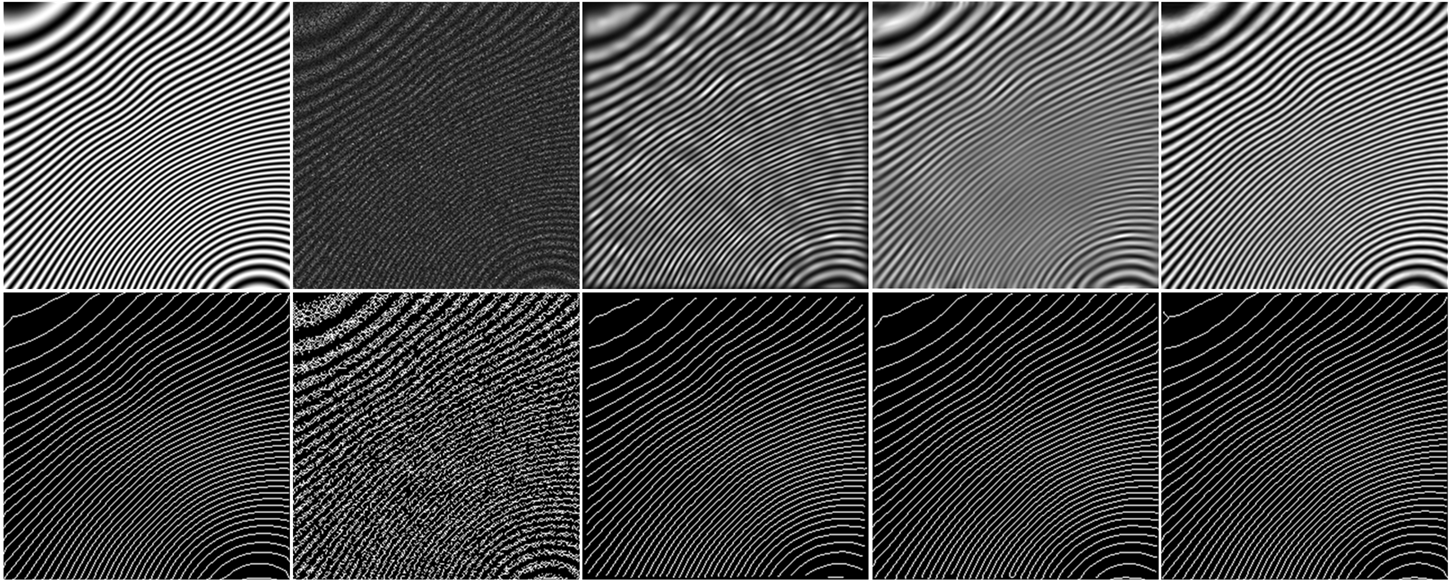}
\caption{Denoising a fine simulated fringe pattern with high-density regions (the 1st row): noise free image, noisy image, WFF, CED, FPD-CNN, respectively. The 2nd row are the corresponding skeletons.}
\label{Fig:7}
\end{figure*}

Then, we perform a test on an experimentally obtained fine optical fringe pattern with high-density regions to verify the generalization ability of the proposed model. As shown in Fig.\ref{Fig:5}, one can see that the result obtained with the ReLU is seriously blurred. However, for the result obtained with the Leaky
ReLU, not only noise is effectively filtered out, but also the geometry of the fringe patten is well preserved and enhanced. The superiority of the Leaky ReLU activation
function can be attributed to its property that it can propagate negative gradients forward, avoiding the problem of so-called ``neuronal death''. So we adopt the Leaky ReLU to enhance generalization performance of proposed CNN model.
\subsection{Results on Simulated Images}
Ten fine simulated images in the testing dataset are are selected to calculate four quantitative evaluation indexes. In favor of the principle of minimizing the MAE, we adjust the parameters of both WFF and CED to be optimal. Average quantification results are shown in the \textbf{Table} \ref{Table:1} with the best in bold. The following analysis illustrates the reasons why the proposed method can achieve significant advantages in quantitative comparisons.
\begin{table}[htbp]
\centering
\captionsetup{justification=centering}
\caption{Average Quantification Results on Ten Simulated images}
\label{Table:1}
\begin{tabular*}{7.5cm}{@{\extracolsep{\fill}}clcccc}
  \toprule
 & Indexs     &  WFF    &  CED       & FPD-CNN &\\
  \midrule
 & PSNR & 15.13   &  12.22     &\bf 27.88 & \\
 & SSIM & 0.684    &  0.554     &\bf 0.972  & \\
 & MAE  & 30.863   &  42.944     &\bf 7.642  & \\
 & Running Time(s)   & 222.17 & 2.29  &\bf 0.57 &\\
  \bottomrule
\end{tabular*}
\end{table}

In Fig.\ref{Fig:6} and Fig.\ref{Fig:7}, we show the denoised results of different methods for two simulated fine fringe pattern with high-density regions. In Fig.\ref{Fig:5}, the results obtained with WFF and CED are not satisfactory, they all produce significant artifacts or blurring. In order not to destroy the signal, WFF adopts a wider frequency band, which also makes partial noise survive in the threshold processing, and consequently reduces the filtering performance. This problem is particularly noticeable when dealing  with speckle noise, as the speckle noise is often superimposed on the frequency spectrum of the fringe pattern, and the simply hard threshold processing cannot completely separate them. The results of the WFF processing in Fig.\ref{Fig:6} reflect the problem just described, not only artifacts and noise still remain in the denoised images, but also existing blurs in high-density regions. The extracted skeleton is also broken in the corresponding region. For CED, its performance is closely related to the accurate estimation of fringe orientation. When the frequency increases gradually, CED can still produce very good results as long as the noise does not distort the structure of fringe pattern, as shown in the 4th column of Fig.\ref{Fig:6}. On the contrary, when the noise has destroyed the structure of the finer regions with high density, the inaccurate calculation of fringe orientation for CED leads to damaged results with details of severe blurring. Consequently, the corresponding skeleton obtained from CED's results are broken severely and have branches.

In Fig.\ref{Fig:7}, it still has fine fringe but with the relatively uniformly high-density fringe distribution compared with Fig.\ref{Fig:6} who has high-variable-density fringe distribution. And the fringe structure is not seriously damaged, so WFF and CED achieved significantly better results than in Fig.\ref{Fig:6}, especially CED method can easily estimate the fringe orientation accurately. But the WFF still generate some artifacts, and CED generated some blurs. Both in Fig.\ref{Fig:6} and Fig.\ref{Fig:7}, the results from WFF and CED is distorted obviously in the image boundary area. Finally, through a large amount of deep learning of the proposed network, an accurate nonlinear mapping from noisy data to noise is approximated by FPD-CNN, which achieves the best results among three compared methods: noise is efficiently filtered off while preserving main features of fringe well, the corresponding skeleton is also quite accurate, although there is a little branch.
\begin{figure*}[h]
\centering
\includegraphics[width=16.1cm,height=5.3cm]{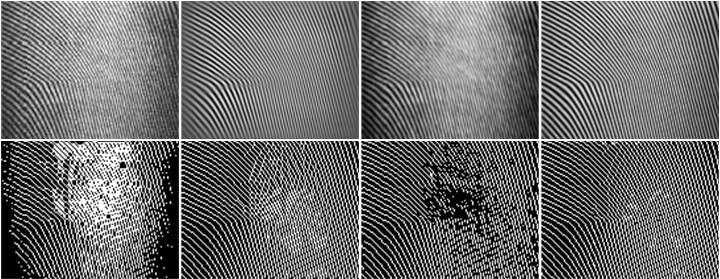}
\caption{Denoising a experimentally obtained fine optical fringe pattern with high-density regions (the 1st row goes from left to right): noisy image, results by WFF, CED, and FPD-CNN, respectively. The 2nd row are the corresponding skeletons.}
\label{Fig:8}
\end{figure*}
\begin{figure*}[h]
\centering
\includegraphics[width=16.1cm,height=5.3cm]{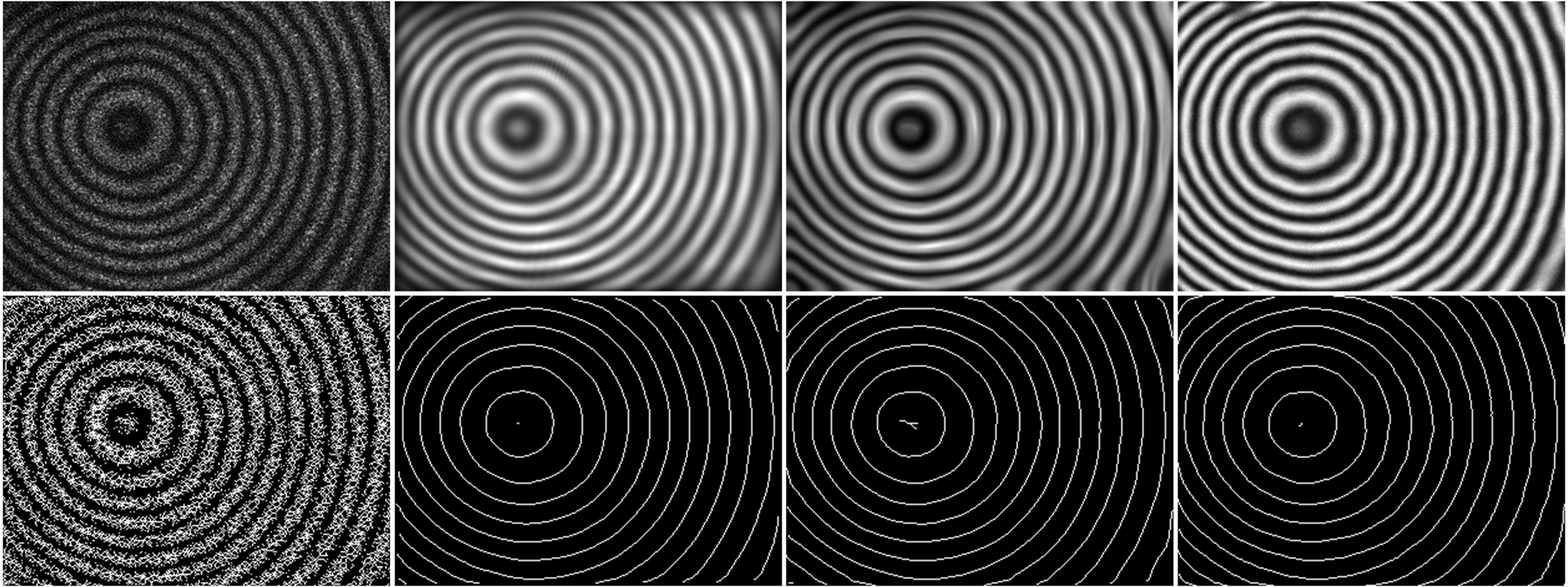}
\caption{Denoising a experimentally obtained fine optical fringe pattern with general-density regions (the 1st row goes from left to right): noisy image, results by WFF, CED, and FPD-CNN, respectively. The 2nd row are the corresponding skeletons.}
\label{Fig:9}
\end{figure*}
\subsection{Results on experimentally obtained Optical Fringe Patterns}
\begin{figure*}[th]
\centering
\includegraphics[width=16.3cm,height=5.55cm]{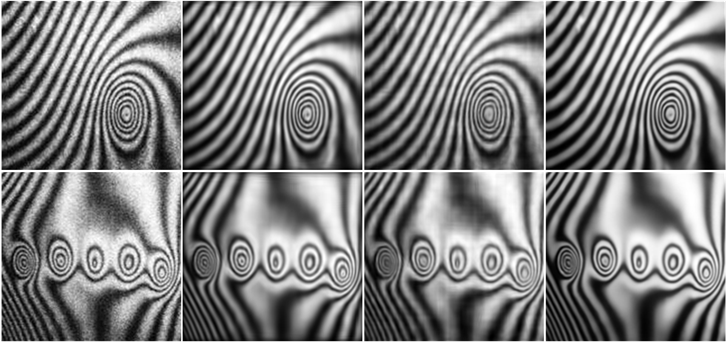}
\caption{Denoising two experimentally obtained fine optical fringe patterns with high-density regions (from left to right): noisy image, results by WFF, CED, and FPD-CNN, respectively.}
\label{Fig:10}
\end{figure*}

\begin{figure*}[th]
\centering
\includegraphics[width=16.3cm,height=5.5cm]{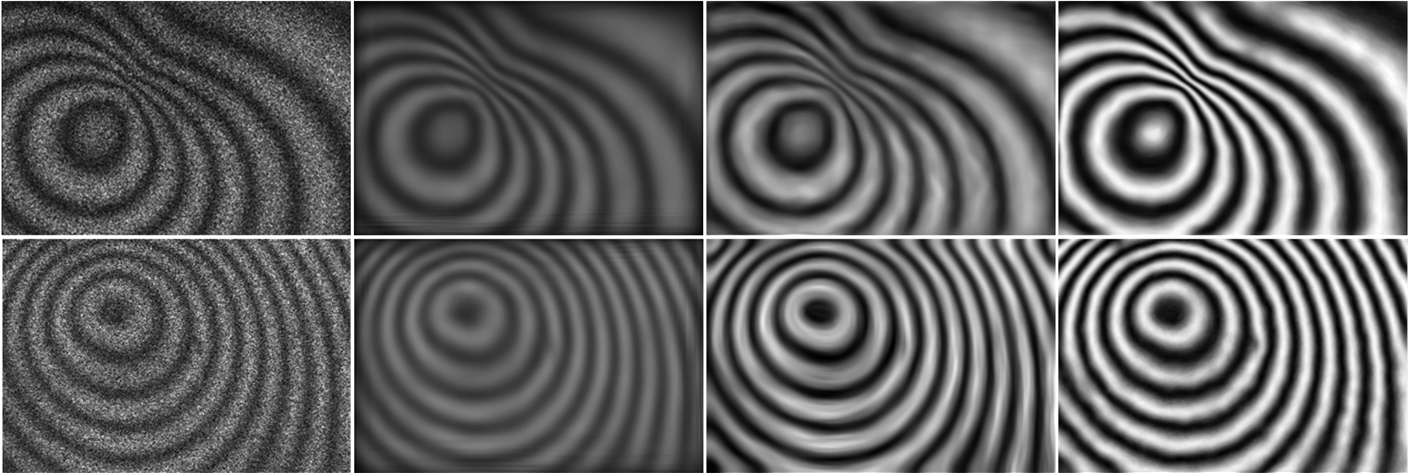}
\caption{Denoising two experimentally obtained optical fringe patterns with general-density regions (from left to right): noisy image, results by WFF, CED, and FPD-CNN, respectively.}
\label{Fig:11}
\end{figure*}

In this section, we test six experimentally obtained  digital interferograms are shown in Fig.\ref{Fig:8}$\sim$Fig.\ref{Fig:11}. In Fig.\ref{Fig:8} and Fig.\ref{Fig:9}
each contain one interferogram, the corresponding denoised results of three methods and the corresponding skeleton line for analysis. The size of these two images are $140\times105$ and $311\times233$, respectively. Fig.\ref{Fig:10} and Fig.\ref{Fig:11} all contain two interferograms with size of $310\times210$, respectively.  We only show the corresponding denoising results in Fig.\ref{eq:10} and Fig.\ref{Fig:11}. In Fig.\ref{Fig:8}, it's a finer and high-density one, the proposed FPD-CNN and WFF have achieved better filtering results preserving fringe structures. Furthermore, the FPD-CNN is superior to WFF in preserving image boundary features, and visually WFF produces a slight blurring result. It is also seen from the extracted skeleton that the proposed method is superior to WFF in preserving fine fringe features. However, CED's results are not satisfactory and obtaining a serious discontinuity in the skeleton. In Fig.\ref{Fig:9} nice filtering results are obtained by three methods for the real fine digital interferograms with general
\begin{table}[h]
\centering
\captionsetup{justification=centering}
\caption{Running time(s) of Fig.\ref{eq:8}$\sim$Fig.\ref{Fig:11} for comparison}
\begin{tabular*}{7.5cm}{@{\extracolsep{\fill}}clccccc}
  \toprule
& Method &Fig.\ref{Fig:8}   &  Fig.\ref{Fig:9}  &  Fig.\ref{Fig:10}  & Fig.\ref{Fig:11} &\\
  \midrule
&   WFF     & 21.71         & 119.37             &  23.12            &     159.64 &       \\
&   CED     & 1.38           & 7.62              &  8.23             &     12.76  &       \\
& FPD-CNN   & 0.21           & 0.60              &  0.46             &     0.58  &       \\
\bottomrule
\end{tabular*}
\label{Table:2}
\end{table}
density, through the boundary geometry of the fringes processed by WFF and CED is obviously not as clear as that of the proposed FPD-CNN. The skeleton corresponding to WFF has a slight loss at the boundary, and in the center of skeleton corresponding to the CED produces some little branches. The edge positions of the skeletons extracted from the filtered images of CED also shift. In Fig.\ref{Fig:10} and Fig.\ref{Fig:11}, they all achieved satisfactory filtering results. In Fig.\ref{Fig:10}, the noise removed by our method is more thorough. And our method and WFF
method are better than CED method in detail preservation. The results obtained by WFF and CED in Fig.\ref{Fig:11} still have distortion phenomena at the boundary, while our method still inevitably has a slight overfitting, albeit better visually. The running time are shown in \textbf{Table} \ref{Table:2}, for Fig.\ref{eq:10} and Fig.\ref{eq:11} are average time. Our method has clearly achieved the fastest processing speed.
\section{Conclusion}
We have proposed a new optical fringe pattern denoising method based on CNNs. Residual learning is used to directly separate noise from noisy image. The FPD-CNN consists of multiple stages with joint training in an end-to-end fashion for improved removal of speckle noise. As an attempt in optical fringe pattern filtering using CNN, our method performs significantly better in terms of preserving main features of fringe with smaller quantitative errors on fine simulated images with high-density regions. The results on experimentally obtained optical fringe patterns further demonstrated that FPD-CNN can deliver perceptually appealing denoising results at a fairly speed. The running time comparisons showed the faster speed of FPD-CNN. Further optimization of the network architecture and the augmentation of corresponding training data based on the pixel statistical distribution of real data are future work.

\section{Acknowledgment}
The research has been supported in part by the National Natural Science Foundation of China (61671276, 11971269); the Natural Science Foundation of Shandong Province of China (ZR2019MF045); the Teaching Reform and Research Project of School of Mathematics of Shandong University.


\end{document}